\documentclass{article}
\usepackage{spconf,amsmath,graphicx,hyperref,cuted,subcaption,booktabs,dblfloatfix,enumitem,bm,booktabs,hyperref,xurl}

\hypersetup{
    colorlinks=true, 
    urlcolor=magenta,   
}

\title{Spiking Vocos: An Energy-Efficient Neural Vocoder\vspace{-0.5em}}

\name{Yukun Chen$^{1}$ \qquad Zhaoxi Mu$^{1}$ \qquad Andong Li$^{2,3}$ \qquad Peilin Li$^{1}$ \qquad Xinyu Yang\thanks{*Corresponding author}$^{1*}$\vspace{-0.5em}}

\address{$^{1}$ Xi'an Jiaotong University, Xi'an, China \\
         $^{2}$ Institute of Acoustics, Chinese Academy of Sciences, Beijing, China \\
         $^{3}$ Chinese Academy of Sciences, Beijing, China\vspace{-0.5em}}

\begin{document}

\let\oldthebibliography\thebibliography
\renewcommand\thebibliography[1]{
  \oldthebibliography{#1}
  \setlength{\parskip}{0.5pt}
  \setlength{\itemsep}{0.5pt plus 0.3ex}
}

\ninept
\maketitle
\begin{abstract}
Despite the remarkable progress in the synthesis speed and fidelity of neural vocoders, their high energy consumption remains a critical barrier to practical deployment on computationally restricted edge devices. Spiking Neural Networks (SNNs), widely recognized for their high energy efficiency due to their event-driven nature, offer a promising solution for low-resource scenarios. In this paper, we propose Spiking Vocos, a novel spiking neural vocoder with ultra-low energy consumption, built upon the efficient Vocos framework. To mitigate the inherent information bottleneck in SNNs, we design a Spiking ConvNeXt module to reduce Multiply-Accumulate (MAC) operations and incorporate an amplitude shortcut path to preserve crucial signal dynamics. Furthermore, to bridge the performance gap with its Artificial Neural Network (ANN) counterpart, we introduce a self-architectural distillation strategy to effectively transfer knowledge. A lightweight Temporal Shift Module is also integrated to enhance the model's ability to fuse information across the temporal dimension with negligible computational overhead. Experiments demonstrate that our model achieves performance comparable to its ANN counterpart, with UTMOS and PESQ scores of $3.74$ and $3.45$ respectively, while consuming only $14.7\%$ of the energy. The source code is available at \url{https://github.com/pymaster17/Spiking-Vocos}.

\end{abstract}
\section{Introduction}
\label{sec:intro}

Vocoding, aiming to restore waveform from acoustic features, is the critical final step of various tasks like audio synthesis, enhancement and conversion. Neural vocoders gradually become mainstream for their improved synthesis quality compared to signal-processing-based counterparts. Although with high quality, auto-regressive vocoders like WaveNet \cite{van2016wavenet}, WaveRNN \cite{kalchbrenner2018efficient} and LPCNet \cite{valin2019lpcnet} suffer from high computational cost and low inference speed. Thus, non-autoregressive GAN-based methods \cite{kumar2019melgan,kong2020hifi,lee2022bigvgan} are developed to output waveform in parallel, significantly improving inference speed and computational efficiency. Diffusion models also represent an important branch of modern vocoder design, characterized by high synthesis fidelity and optimized inference speed \cite{nguyen2024fregrad,luo2025wavefm}.\par

Despite the success of time-domain vocoders, they all need computationally-intensive upsample layers to generate the waveform at sample point level, without leveraging the high efficiency of the inverse Short-Time Fourier Transform (iSTFT) for upsampling. Frequency-domain vocoders, aim to generate Fourier spectral coefficients, which can be reconstructed to waveform by iSTFT losslessly. Compared with their time-domain counterparts, frequency-domain vocoders have more lightweight structures in nature, without the burden to generate long waveform directly. iSTFTNet \cite{kaneko2022istftnet} designs a hybrid network structure based on HiFiGAN \cite{kong2020hifi}, replacing the last few upsampling layers with iSTFT. Vocos \cite{siuzdak2023vocos} adopts a consistent structure without any upsampling layers, maintaining the same temporal resolution along all layers. Benefiting from its lightweight structure, Vocos achieves a Real-Time Factor (RTF) several times higher than that of iSTFTNet. APNet2 \cite{du2023apnet2} explicitly models phase spectrum with a proposed anti-wrapping function, improve the accuracy of phase prediction. RFWave \cite{liu2024rfwave} estimates sub-band of complex spectrograms individually with rectified flow, with an overlap loss to reduce inconsistencies among them.\par

Although frequency-domain vocoders like Vocos \cite{siuzdak2023vocos} improve computational efficiency, they are not explicitly optimized for power consumption. This paper addresses this gap by leveraging Spiking Neural Networks (SNNs), a bio-inspired computing paradigm known for its exceptional energy efficiency due to its event-driven, accumulation-based (AC) operations \cite{maass1997networks}. Training deep SNNs is challenging due to the non-differentiable nature of spike generation. While ANN-to-SNN conversion methods exist \cite{bu2023optimal}, direct training using a surrogate gradient \cite{neftci2019surrogate,zhou2024direct} has become the mainstream approach for achieving low-latency, high-performance models. However, directly substituting Artificial Neural Networks (ANNs) with SNNs typically results in a performance drop due to inherent challenges like the information bottleneck from binary spikes and suboptimal temporal modeling \cite{cao2024spiking,wang2024spikevoice}. To close the performance gap between ANNs and SNNs, several techniques have been proposed. Knowledge Distillation (KD) has proven effective for transferring knowledge from a pre-trained ANN teacher to an SNN student by matching intermediate features or final outputs \cite{qiu2024self, yang2025efficient}. Concurrently, methods for improving temporal processing have been explored. The Temporal Shift Module (TSM) \cite{yu2025ts} is a lightweight yet effective technique that shifts feature channels across the time dimension to fuse past, present, and future information with negligible computational overhead. In this paper, we propose Spiking Vocos, the first high-fidelity, energy-efficient SNN-based vocoder. Integrated with both KD and TSM, Spiking Vocos synergistically addresses the performance limitations of SNNs in the context of audio generation.\par

Our contributions are:
\begin{itemize}[itemsep=0.1em, topsep=0.1em]
\item We are the first to introduce an SNN into a frequency-domain vocoder, designing an efficient spiking ConvNeXt module that significantly reduces energy consumption while maintaining high perceptual quality.
\item We propose a self-architectural distillation framework tailored for vocoding, which effectively boosts the synthesis quality of the SNN model.
\item We validate the effectiveness of the Temporal Shift Module in the audio synthesis domain, demonstrating its capacity to enhance the temporal processing of SNNs.
\end{itemize}

\begin{figure*}[htb]
    \centering

    % 第一个子图 (a)
    \begin{subfigure}[b]{0.58\textwidth}
        \centering
        \includegraphics[width=\textwidth]{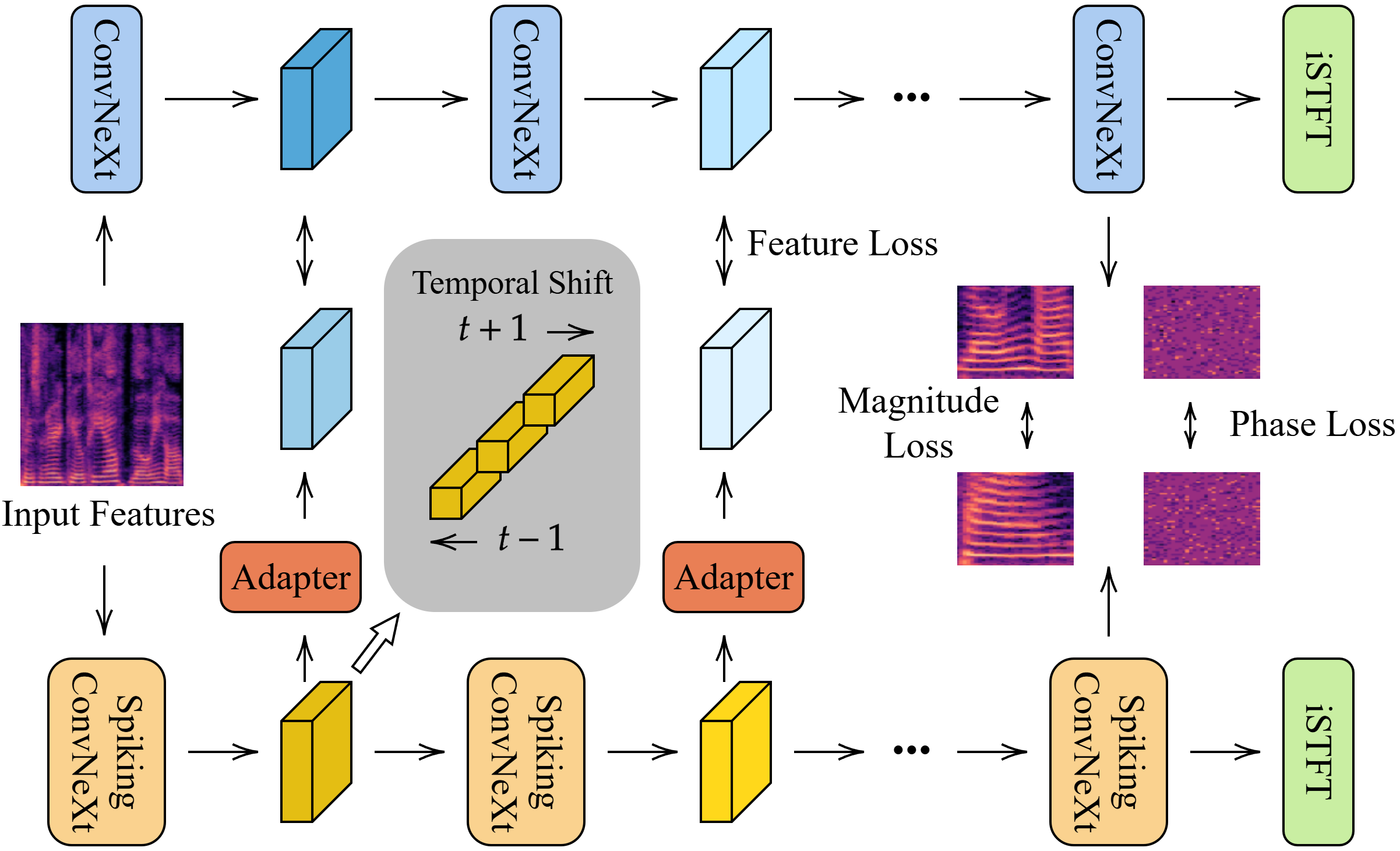}
        \caption{}
        \label{fig:sub-a} % 更改标签，以便引用子图
    \end{subfigure}
    \hfill
    % 第二个子图 (b)
    \begin{subfigure}[b]{0.40\textwidth}
        \centering
        \includegraphics[width=\textwidth]{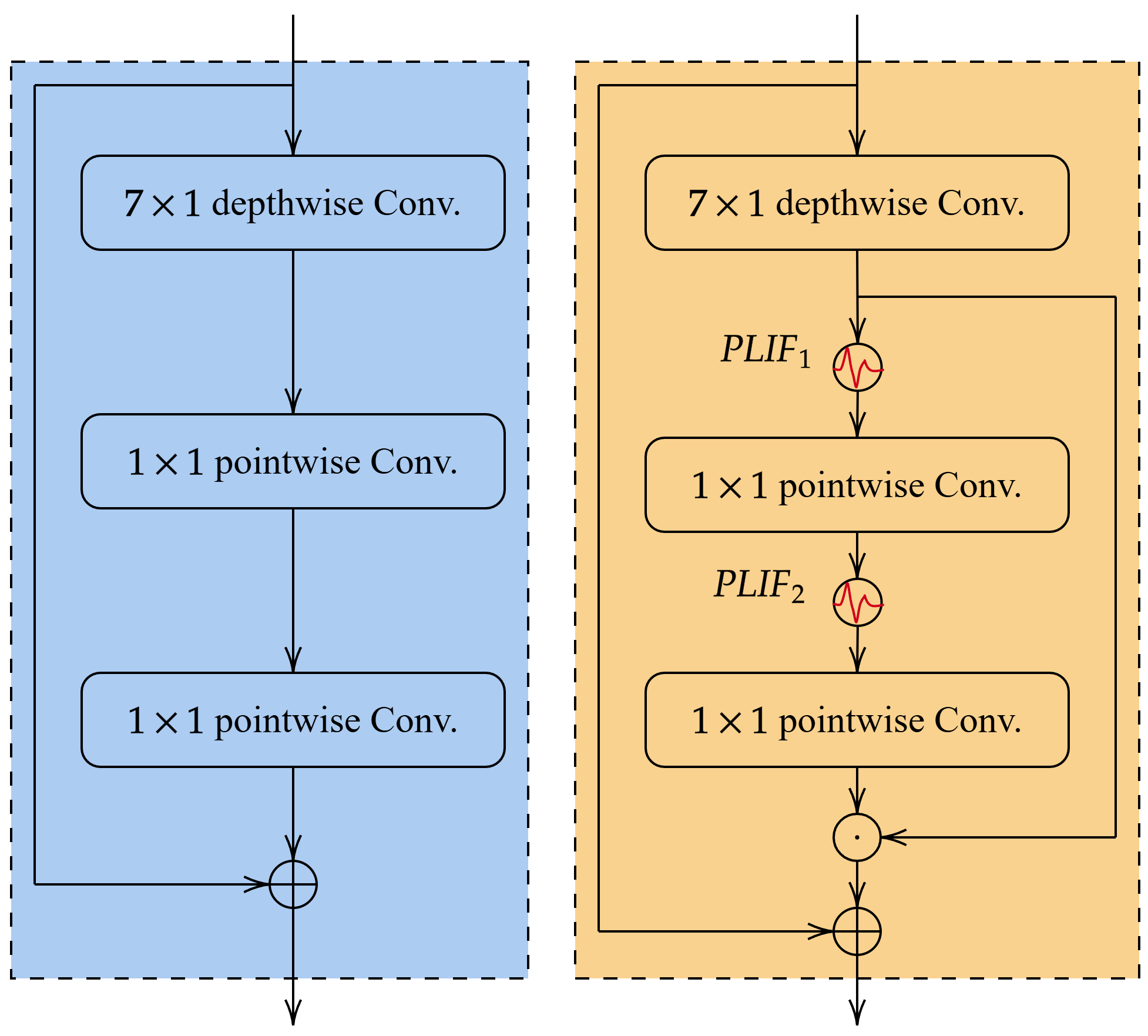}
        \caption{}
        \label{fig:sub-b} % 更改标签，以便引用子图
    \end{subfigure}

    % 整个图组的大标题
    \caption{(a) The overall architecture of the Spiking Vocos generator. The input mel-spectrogram is processed by a stack of Spiking ConvNeXt blocks, where the Temporal Shift Module (TSM) is applied in each block. (b) A comparison between the standard ConvNeXt block (left) and our proposed Spiking ConvNeXt block (right). Our design introduces two PLIF neurons before the computationally intensive pointwise convolutions and adds an amplitude shortcut path to mitigate the information bottleneck.}
    \label{fig:main-fig} % 整个图组的标签
\end{figure*}

\section{Method}
\label{sec:method}

This section details the proposed Spiking Vocos, an ultra-low-power vocoder that adapts the high-efficiency Vocos framework to the spiking domain. An overview of the model architecture is presented in Fig.~\ref{fig:sub-a}. We first introduce the core Spiking ConvNeXt block in Section~\ref{sec:spiking convnext}, which forms the backbone of our generator. Next, Section~\ref{sec:self-architectural distillation} elaborates on the self-architectural distillation paradigm used to bridge the performance gap between the ANN and SNN models. Finally, Section~\ref{sec:temporal shift module} describes the integration of the Temporal Shift Module to enhance the model's temporal modeling capabilities.

\subsection{Spiking ConvNeXt Block}
\label{sec:spiking convnext}

The fundamental building block of the Spiking Vocos generator is the Spiking ConvNeXt block, which is adapted from the standard ConvNeXt architecture \cite{liu2022convnet}. As illustrated in Fig.~\ref{fig:sub-b}, our design prioritizes computational efficiency. Since the two pointwise convolutions account for the majority of the computational load, we choose to insert spiking neurons directly before them. This ensures that these computationally intensive operations are performed on sparse, binary spikes, maximizing the energy savings of the SNN.\par

For the neuronal model, we employ the Parametric Leaky Integrate-and-Fire (PLIF) neuron \cite{fang2021incorporating} for a higher diversity and expressiveness. Unlike the standard LIF neuron \cite{maass1997networks}, the PLIF neuron features a learnable time constant $\tau$, which allows it to adaptively balance the influence of present input against past memory. The dynamics of the PLIF neuron follow a three-stage process at each timestep $t$: charging, firing, and resetting, which can be described as:

\vspace{-1em}
\begin{align}
H_{t} & =V_{t-1} +\frac{1}{\tau }\left( X_{t} -\left( V_{t-1} -V^{\text{re}}\right)\right) \label{eq:Ht} \\
S_{t} & =\Theta \left( H_{t} -V^{\text{th}}\right) \label{eq:St} \\
V_{t} & =V^{\text{re}} S_{t} +H_{t}( 1-S_{t}) \label{eq:Vt}
\end{align}

Here, Eq. \eqref{eq:Ht} describes the charging step, where the membrane potential $V_{t-1}$ from the previous timestep is updated with the input current $X_t$ to produce the new potential $H_t$. Eq. \eqref{eq:St} represents the firing mechanism, where $\Theta(\cdot)$ is the Heaviside step function that produces an output spike $S_{t} \in \{0,1\}$ if $H_t$ exceeds the firing threshold $V^{\text{th}}$. Finally, Eq. \eqref{eq:Vt} is the resetting function, where the membrane potential $V_t$ is reset to $V^{\text{re}}$ if a spike was fired.\par

A critical challenge in SNNs is the information bottleneck caused by the all-or-none nature of spiking. As shown in Eq. \eqref{eq:St}, all supra-threshold inputs are mapped to a spike firing, effectively erasing crucial amplitude information. This "saturation phenomenon" can degrade the final performance. To address this, we introduce an amplitude shortcut path to circumvent the bottleneck:

\vspace{-0.5em}
\begin{equation}
    Z_{\text{recover}}=\left| Z_{\text{in}} \right| \odot Z_{\text{out}},
\end{equation}

where $\odot$ denotes element-wise multiplication. This operation re-injects the amplitude information into the data stream (Fig.~\ref{fig:sub-b}), allowing the model to benefit from the computational sparsity of spikes without sacrificing essential signal dynamics.

\subsection{Self-architectural Distillation}
\label{sec:self-architectural distillation}

\begin{table*}[ht]
 \centering
 \caption{Performance comparison of the baseline ANN Vocos and Spiking Vocos variants on the LibriTTS test-clean set.}
 \label{tab:spiking}
 \begin{tabular}{lccccc}
  \toprule
  \textbf{Model} & \textbf{UTMOS ($\uparrow$)} & \textbf{PESQ ($\uparrow$)} & \textbf{ViSQOL ($\uparrow$)} & \textbf{V/UV F$1$ ($\uparrow$)} & \textbf{Periodicity ($\downarrow$)} \\
  \midrule
  Vocos (ANN Baseline) & $3.82$ & $3.65$ & $4.67$ & $0.9600$ & $0.108$ \\
  \midrule
  Spiking Vocos ($8$-step) & $3.80$ & $3.49$ & $4.66$ & $0.9566$ & $0.114$ \\
  \midrule
  Spiking Vocos ($4$-step) & $3.46$ {\footnotesize ($-0.36$)} & $3.31$ {\footnotesize ($-0.34$)} & $4.63$ {\footnotesize ($-0.04$)} & $0.9522$ {\footnotesize ($-0.0078$)} & $0.127$ {\footnotesize ($+0.019$)} \\
  \quad + TSM & $3.71$ {\footnotesize ($-0.11$)} & $3.36$ {\footnotesize ($-0.29$)} & $4.65$ {\footnotesize (\bm{$-0.02$})} & $0.9539$ {\footnotesize ($-0.0061$)} & $0.116$ {\footnotesize (\bm{$+0.008$})} \\
  \quad + Distillation & $3.70$ {\footnotesize ($-0.12$)} & $3.43$ {\footnotesize ($-0.22$)} & $4.65$ {\footnotesize (\bm{$-0.02$})} & $0.9559$ {\footnotesize (\bm{$-0.0041$})} & $0.118$ {\footnotesize ($+0.010$)} \\
  \quad + TSM \& Distillation & $3.74$ {\footnotesize (\bm{$-0.08$})} & $3.45$ {\footnotesize (\bm{$-0.20$})} & $4.65$ {\footnotesize (\bm{$-0.02$})} & $0.9558$ {\footnotesize ($-0.0042$)} & $0.116$ {\footnotesize (\bm{$+0.008$})} \\
  \bottomrule
 \end{tabular}
\end{table*}

While the surrogate gradient method enables direct SNN training, challenges such as training instability and a persistent performance gap compared to ANNs remain \cite{neftci2019surrogate, guo2022recdis}. To address this, we employ a self-architectural knowledge distillation (KD) framework. As its ANN counterpart, Vocos serves as an ideal teacher for Spiking Vocos due to their identical macro-architectures. Our distillation strategy provides guidance at two critical levels: intermediate feature representations and final spectral outputs.\par

To align the internal representations of the two models, we distill knowledge layer-wise. Lightweight adapters, each consisting of a linear layer and an activation function, are employed to project the student's intermediate features ($z_{\text{stu}}$) into the teacher's feature space. The alignment is enforced by minimizing the Mean Squared Error (MSE) over all $N$ distilled blocks between the projected student features and the teacher's features ($z_{\text{tea}}$):

\vspace{-0.5em}
\begin{equation}
    \mathcal{L}_{\text{feat}} =\sum_{n=1}^N\Vert F( z_{\text{stu}}) -z_{\text{tea}}\Vert _{2}^2,\label{eq:feature}
\end{equation}

where $z$ denotes an intermediate representation, and $F$ is the adapter's projection function. This process encourages the student SNN to mimic the layer-wise behavior of the teacher.\par

At the final layer, we apply distinct distillation objectives for the magnitude and phase spectra to account for their different properties. The magnitude loss, $\mathcal{L}_M$ is the L1 distance between the logarithmic magnitudes of the student and teacher models:

\vspace{-0.5em}
\begin{equation}
\mathcal{L}_{\text{M}} =\Vert \log( A_{\text{stu}}) -\log( A_{\text{tea}})\Vert _{1},\label{eq:mag}
\end{equation}

Distilling the phase spectrum is more challenging due to its periodic nature, which causes wrapping around $\pm\pi$. Inspired by \cite{du2023apnet2}, we apply an anti-wrapping function $f_{AW}$ to the phase difference, which maps the error to its principal value:

\begin{equation}
    f_{\text{AW}}(x) = \left| x - 2 \pi \cdot \text{round}\left(\frac{x}{2 \pi}\right) \right| \label{eq:anti}
\end{equation}

The total phase loss, $\mathcal{L}_{\text{P}}$, is a composite of three components that capture different aspects of phase correctness: instantaneous phase loss ($\mathcal{L}_{\text{IP}}$), group delay loss ($\mathcal{L}_{\text{GD}}$), and phase time difference loss ($\mathcal{L}_{\text{PTD}}$):

\vspace{-1.5em}
\begin{align}
\mathcal{L}_{\text{IP}} & =\operatorname{mean}[ f_{AW}( \phi _{\text{tea}} -\phi _{\text{stu}})]\\
\mathcal{L}_{\text{GD}} & =\operatorname{mean}[ f_{AW}( \nabla _{\omega } \phi _{\text{tea}} -\nabla _{\omega } \phi _{\text{stu}})]\\
\mathcal{L}_{\text{PTD}} & =\operatorname{mean}[ f_{AW}( \nabla _{t} \phi _{\text{tea}} -\nabla _{t} \phi _{\text{stu}})]\\
\mathcal{L}_{\text{P}} & =\mathcal{L}_{\text{IP}} +\mathcal{L}_{\text{GD}} +\mathcal{L}_{\text{PTD}}
\end{align}

where $\nabla _{\omega}$ and $\nabla _{t}$ denote the derivatives with respect to frequency and time. The cooperation of the three phase losses can constrain the prediction accuracy at point level as well as the consistency in time and frequency dimension.\par

The total knowledge distillation loss is a weighted sum of these components:

\vspace{-1em}
\begin{equation}
    \mathcal{L}_{\text{KD}} = \lambda_{\text{feat}}\mathcal{L}_{\text{feat}} + \lambda_{\text{P}}\mathcal{L}_{\text{P}} + \lambda_{\text{M}}\mathcal{L}_{\text{M}}.\label{KD}
\end{equation}

This multi-faceted loss function provides comprehensive guidance, encouraging the SNN to replicate not only the final output but also the internal computational steps of its high-performing ANN counterpart.

\subsection{Temporal Shift Module Integration}
\label{sec:temporal shift module}

SNN has a similar temporal dynamics to RNN, where an implicit state (membrane potential) is passed from one block to the next. Moreover, the inherent causal nature of SNN causes its blindness to future timesteps, termed as "partial-time dependency" \cite{wang2024spikevoice}. Temporal Shift Module \cite{yu2025ts} is designed to allow every block explicitly "see" the information from past and future simultaneously.\par

As shown in Fig.~\ref{fig:sub-a}, the tensor of intermediate feature $Z_{\text{org}}$ is split into three parts by channel indexs $C_{-1}, C_{0}, C_{1}$, where $C_{-1}<C_{0}<C_{1}$. The three channel groups will be shifted $-1, 0, 1$ timestep respectively, with proper padding and truncation to maintain consistent shape:

\vspace{-1em}
\begin{equation}
Z_{\text{shift}}[ t,\ c,\ ...] =\begin{cases}
Z_{\text{org}}[ t+1,\ c ,\ ...] & 0 \leq c< C_{-1}\\
Z_{\text{org}}[ t,\ c ,\ ...] & C_{-1} \leq c< C_{0}\\
Z_{\text{org}}[ t-1,\ c ,\ ...] & C_{0} \leq c\leq C_{1}
\end{cases}
\end{equation}

However, shifting channels risks diluting information from the original timestep. Therefore, a residual connection is adopted to combine the original features with the shifted ones:

\vspace{-0.5em}
\begin{equation}
    Z=\alpha \odot Z_{\text{shift}}+Z_{\text{org}},
\end{equation}

where $\alpha$ is a hyperparameter controlling the intensity of the temporal shift.

\section{Experiments}
\label{sec:typestyle}

\subsection{Experimental Setup}
\label{sec:setup}

We train spiking models on the complete training set of LibriTTS \cite{zen2019libritts}. The ANN-based Vocos model \cite{siuzdak2023vocos} is trained as baseline, as well as the teacher model for distillation. The original $24$ kHz audio is compressed into 100-dimension mel-scaled spectrograms with $n_{fft}=1024$ and $n_{hop}=256$. All Spiking Vocos variants and the baseline are trained for 1 million generator and discriminator steps. We use the AdamW optimizer with $\beta_1=0.9$ and $\beta_2=0.999$.\par

For the TSM, we use a fixed channel split ($C_{-1}=\frac{1}{4}C_{1}, C_{0}=\frac{3}{4}C_{1}$) for training stability, with a residual weight $\alpha=0.5$. A crucial implementation detail for models using both TSM and distillation is that the intermediate distillation points are shifted to the subsequent ConvNeXt block. This modification prevents the feature alignment objective from being disrupted by the temporal shift operation.\par

All models are evaluated on the test-clean subset of LibriTTS. For objective metrics, we use UTMOS \cite{saeki2022utmos}, a pseudo MOS metric correlated well with human Mean Opinion Scores (MOS). We also employ acoustic metrics including PESQ \cite{rix2001perceptual} and ViSQOL \cite{chinen2020visqol} for the evaluation of signal quality. Following standard practice \cite{morrison2021chunked}, we measure objective characteristics using the F1 score for voiced/unvoiced classification (V/UV F1) and periodicity error. At last, a subjective listening test is conducted as the gold standard for the synthesis quality. A pretrained HiFiGAN\footnote{\url{https://huggingface.co/speechbrain/tts-hifigan-libritts-22050Hz}} on LibriTTS is used as the baseline of time-domain vocoder.\par

\begin{table}[h]
 \centering
 \caption{Subjective evaluation metrics – 5-scale Mean Opinion Score (MOS) and Similarity Mean Opinion Score (SMOS) with 95\% confidence interval.}
 \label{tab:MOS}
 \setlength{\tabcolsep}{15pt} 
 \begin{tabular}{lcc}
  \toprule
  \textbf{Model} & \textbf{MOS ($\uparrow$)} & \textbf{SMOS ($\uparrow$)} \\
  \midrule
  Groud truth & $3.92 \pm 0.14$ & $4.14 \pm 0.12$ \\
  \midrule
  Vocos & $3.80 \pm 0.14$ & $3.79 \pm 0.12$ \\
  HiFiGAN & $3.61 \pm 0.13$ & $3.72 \pm 0.12$ \\
  \midrule
  Spiking Vocos & $3.69 \pm 0.13$ & $3.69 \pm 0.13$ \\
  \bottomrule
 \end{tabular}
 \setlength{\tabcolsep}{6pt}
\end{table}

\subsection{Audio Quality Evaluation}
\label{sec:audio quality}

The audio quality evaluation results are presented in Table~\ref{tab:spiking}. The baseline ANN Vocos sets a strong benchmark with a UTMOS of $3.82$. As expected, the vanilla 4-step Spiking Vocos exhibits a significant performance degradation, highlighting the challenge of direct SNN implementation. Increasing the simulation to 8 timesteps substantially closes this gap, achieving a UTMOS of $3.80$, nearly matching the ANN. This demonstrates the feasibility of high-quality spiking vocoders, but at the cost of doubled computational latency.\par

Focusing on the more efficient 4-timestep setting, our ablation studies validate the effectiveness of the proposed techniques. Integrating the Temporal Shift Module (TSM) alone provides a dramatic improvement, boosting the UTMOS from $3.46$ to $3.71$. This confirms that enhancing temporal information fusion is critical for SNN-based audio synthesis. Similarly, applying self-architectural distillation yields a comparable UTMOS gain ($3.70$) and provides the largest improvement in the PESQ score among the ablations, indicating its success in transferring the teacher's fine-grained spectral knowledge.\par

Crucially, when combining both TSM and distillation, our 4-step Spiking Vocos achieves the best SNN performance with a UTMOS of $3.74$ and a PESQ of $3.45$. While the perceptual quality is high, a notable gap remains in the PESQ score compared to the ANN baseline. We hypothesize this is due to the inherent quantization effect of binary spikes, which may slightly reduce the reconstruction precision of the complex spectrum. Although this impacts metrics like PESQ that rely on signal-level consistency, subjective evaluations (Table \ref{tab:MOS}) suggest that human perception is more robust to this type of error. This result demonstrates that our methods work synergistically to bridge the performance gap, achieving high perceptual fidelity with only 4 timesteps.

\subsection{Energy Consumption Analysis}

\begin{figure}[ht]
    \centering
    \includegraphics[width=0.85\columnwidth]{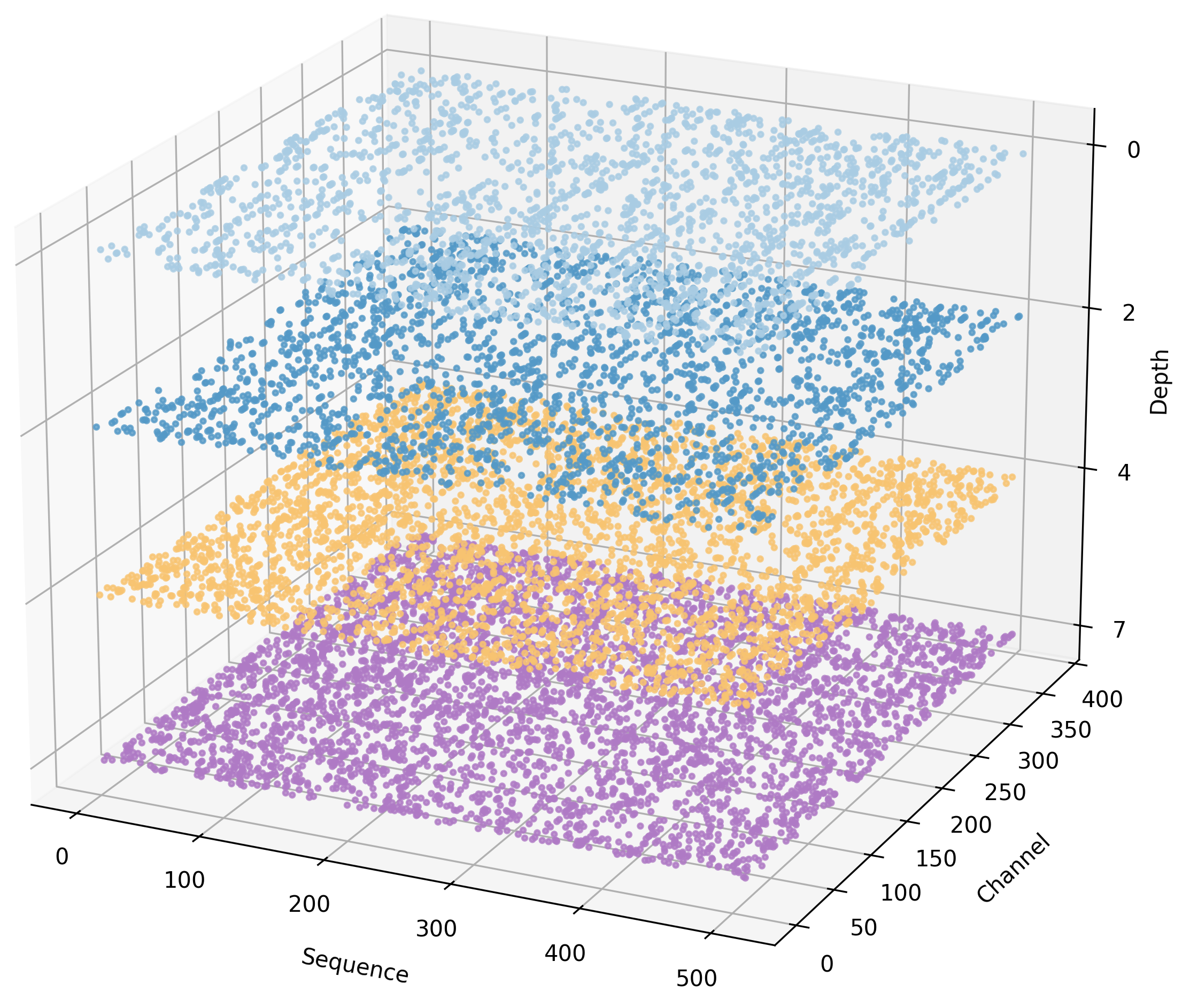}
    \caption{Visualization of spike activity at different depths.}
    \label{fig:spike_visualization}
\end{figure}

The primary motivation for using SNNs is their superior energy efficiency, which stems from their event-driven nature. Fig.~\ref{fig:spike_visualization} visualizes the sparse spike activity in our model, where each dot represents a firing event. The firing rate increases with network depth, a pattern also observed in other SNN audio models like SpikeVoice \cite{wang2024spikevoice}. As shown in Table \ref{tab:energy}, enabling TSM and distillation moderately increases the average firing rate. This suggests a trade-off, where a slight increase in neuronal activity is a necessary cost for achieving higher spectral reconstruction accuracy.\par

\begin{table}[h]
 \centering
 \caption{Estimated theoretical energy consumption of Spiking Vocos variants and the baseline when $L=1000$.}
 \label{tab:energy}
 \begin{tabular}{lcc}
  \toprule
  \textbf{Model} & \textbf{Firing Rate} & \textbf{Energy Con. (pJ)} \\
  \midrule
  Vocos & / & $58.0 \times 10^9$ \\
  \midrule
  Spiking Vocos ($8$-step) & $14.7\%$ & $14.4 \times 10^9$ \\
  \midrule
  Spiking Vocos ($4$-step) & $12.9\%$ & $6.4 \times 10^9$ \\
  \quad + TSM & $14.1\%$ & $6.9 \times 10^9$ \\
  \quad + Distillation & $18.0\%$ & $8.7 \times 10^9$ \\
  \quad + TSM \& Distillation & $17.6\%$ & $8.5 \times 10^9$ \\
  \bottomrule
 \end{tabular}
\end{table}

The energy consumption of the Spiking ConvNeXt block is dominated by its convolution operations. While the depthwise convolution still requires continuous-valued MACs, the computationally-intensive pointwise convolutions now operate on sparse, binary spike inputs, converting most MACs to energy-efficient ACs. The energy can be modeled as:

\vspace{-1em}
\begin{align}
    E_{\text{dwConv}} & = K_{d} \cdot C_{\text{in}} \cdot L \cdot T \cdot E_{\text{MAC}}, \label{eq:dw}\\
    E_{\text{pwConv}} & = K_{p} \cdot C_{\text{in}} \cdot C_{\text{out}} \cdot L \cdot T \cdot r \cdot E_{\text{AC}}, \label{eq:pw}
\end{align}

where $K$ is kernel size, $C$ is channel count, $L$ is sequence length, and $T$ is the SNN timestep. Table \ref{tab:energy} presents the theoretical energy consumption based on established costs for 32-bit floating-point AC ($E_{\text{AC}} \approx 0.9$ pJ) and MAC ($E_{\text{MAC}} \approx 4.6$ pJ) operations on 45nm technology \cite{xing2024spikelm}. Our final 4-step Spiking Vocos model, with an average firing rate of $r=17.6\%$, is estimated to consume only $14.7\%$ of the energy of the ANN-based Vocos. This represents a greater than 6.8x improvement in energy efficiency, highlighting the practical benefits of our approach.\par

\section{Conclusion}
\label{sec:conclusion}

In this work, we introduced Spiking Vocos, the first SNN-based frequency-domain vocoder designed for high-fidelity and ultra-low-power audio synthesis. We addressed the core challenges of applying SNNs to audio generation by designing a Spiking ConvNeXt block with an amplitude shortcut to prevent information loss. To bridge the performance gap with the original ANN model, we employed a self-architectural distillation framework tailored for vocoding and integrated a Temporal Shift Module to enhance temporal modeling. Our experiments demonstrate that the proposed 4-timestep model achieves perceptual quality comparable to the baseline ANN Vocos, while consuming merely $14.7\%$ of the energy.

\vfill\pagebreak

\bibliographystyle{IEEEbib}
\bibliography{refs}

\end{document}